%% file: main.tex
\documentclass[journal]{IEEEtran}

\input{mycommands}

\usepackage{amsmath,amsfonts}
\usepackage{algorithm}
\usepackage{array}
\usepackage{booktabs}
\usepackage{multirow}
\usepackage{textcomp}
\usepackage{stfloats}
\usepackage{url}
\usepackage{verbatim}
\usepackage{graphicx}
\usepackage{cite}
\usepackage{adjustbox}
\usepackage{algpseudocode}
\usepackage{pgfplots}
\usepackage{xcolor}
\usepackage{tikz}
\usepackage{siunitx}
\pgfplotsset{compat=1.18}
\usepackage{pgfplotstable}
\usepgfplotslibrary{groupplots}
\usepackage[caption=false,font=footnotesize]{subfig}
\begin{document}

\title{\name: Distributed Inference for Foundation Models at Edge}

\author{Muhammad Azlan Qazi, Alexandros Iosifidis, and Qi Zhang
\thanks{Muhammad Azlan Qazi and Qi Zhang are with the Department of Electrical and Computer Engineering, Aarhus University, Denmark (e-mail: maq@ece.au.dk; qz@ece.au.dk)}
\thanks{Alexandros Iosifidis is with the Faculty of Information Technology
 and Communication Sciences, Tampere University, Finland (e-mail: alexandros.iosifidis@tuni.fi)}
}

\markboth{Journal of \LaTeX\ Class Files,~Vol.~XX, No.~X, XXX~XXXX}%
{Shell \MakeLowercase{\textit{et al.}}: A Article Using IEEEtran.cls for IEEE Journals}


\maketitle

\begin{abstract}

Foundation models (FMs) have achieved remarkable success across a wide range of applications, from image classification to natural langurage processing, but pose significant challenges for deployment at edge. This has sparked growing interest in developing practical and efficient strategies for bringing foundation models to edge environments. 
In this work, we propose \name, a communication-efficient and compute-aware strategy for distributed Transformer inference on edge devices. Our method leverages a \sm representation to approximate intermediate output features, drastically reducing inter-device communication. Additionally, we restructure the self-attention mechanism to eliminate redundant computations caused by per-device Key/Value calculation in position-wise partitioning and design a partition-aware causal masking scheme tailored for autoregressive models. We evaluate \name on ViT, BERT, and GPT-2 across diverse datasets, namely CIFAR-10, CIFAR-100, ImageNet-1k, GLUE, and CBT. Our results demonstrate substantial reductions in communication overhead (up to 99.2\% for BERT at compression rate $\textit{CR} = 128$) and per-device computation (51.24\% for BERT at the same setting), with only minor accuracy degradation. This method offers a scalable and practical solution for deploying foundation models in distributed resource-constrained environments.

\end{abstract}


\section{Introduction}

\IEEEPARstart{T}{he} introduction of the Transformer architecture has revolutionized the field of artificial intelligence, leading to groundbreaking advancements across natural language processing (NLP), computer vision, and multimodal learning. Foundation models (FMs) such as GPT ~\cite{radford2019language, brown2020language}, LLaMA~\cite{touvron2023llama,touvron2023llama2}, BERT~\cite{devlin2019bert} and ViT~\cite{dosovitskiy2021imageworth16x16words}, DeepSeek’s R1~\cite{guo2025deepseek}, and Veo-3 have significantly transformed how people interact with AI, becoming integral to everyday applications, from document summarization and dialogue systems to image and video generation.

However, the extraordinary capabilities of these models come at the cost of enormous model size. These FMs typically contain hundreds of billions of parameters, which are essential for capturing complex patterns and long-range dependencies in the input data. As a result, they require high-performance accelerators (e.g., GPUs with large memory capacity), which are typically hosted on cloud servers. This cloud-based deployment introduces several limitations: suffering from large and time-varying communication latency, imposing high communication network traffic, and ongoing maintenance costs. More importantly, it raises critical privacy concerns, particularly when sensitive user data is transmitted to and processed by remote servers~\cite{GILL2024100116}.

Edge computing has emerged as a promising alternative to mitigate these issues~\cite{10767295}. By enabling model inference closer to the data source, edge computing reduces latency and improves data privacy. However, deploying large FMs directly on edge device remains infeasible due to their limited compute power and memory resources. To address this, some recent works proposed offloading heavy layers to the cloud while running lighter computations on the edge~\cite{arzovs2024distributed,ALI20252586,10994362}. These edge-cloud continuum strategies, suffer from inconsistent latency and continued reliance on cloud connectivity. Others have explored quantization to compress the model footprint, but often at the cost of degraded performance~\cite{Shao2023OmniQuantOC,MLSYS2024_42a452cb,Shen_Dong_Lu_Kong_Li_Lin_Wu_Wang_2024}.
\vspace{0.2em}

\begin{figure}
\centering
\includegraphics[width=\columnwidth]{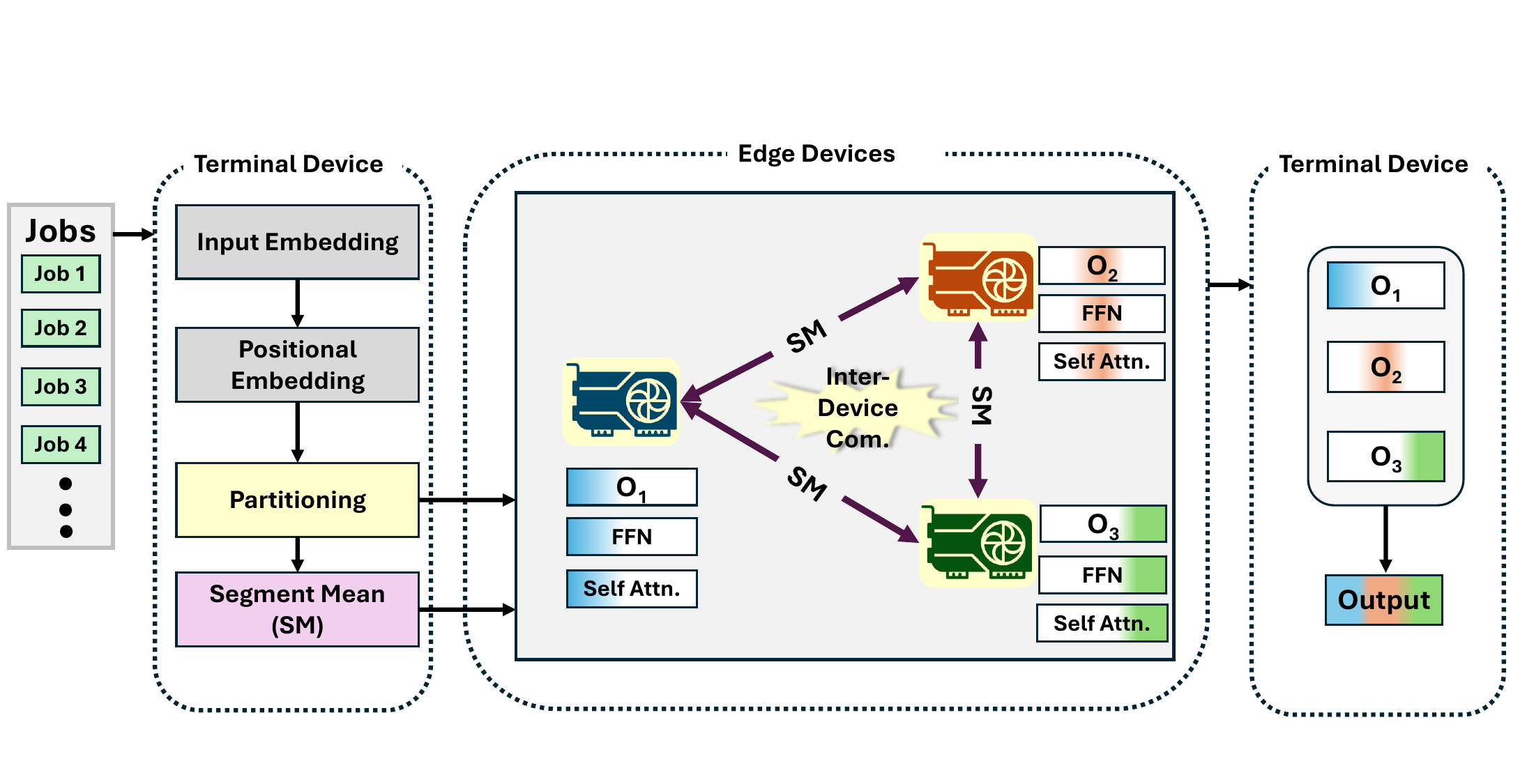}
\caption{Overview of the proposed model architecture. Input tokens are embedded and partitioned on the terminal device, with each partition summarized using \sm and distributed to edge devices. Each edge device performs local self-attention and feedforward network computation, computes its own \sm, and exchanges it with other devices. This process is repeated across Transformer blocks. Finally, output partitions are aggregated at the terminal device to produce the final result.}
\label{arch}
\end{figure}

Distributed inference for large Transformer models relies on several parallelization strategies to balance computational load and communication overhead. In \textit{data parallelism}~\cite{10.1145/79173.79181} the same model is replicated across devices and each device processes a different subset of the input data. However, this approach is less effective for inference tasks, particularly when batch sizes are small. In contrast, \textit{model parallelism}~\cite{NIPS2012_6aca9700} distributes model parameters across devices, allowing deployment of models that exceed single-device memory limits. While this reduces per-device memory requirements, it introduces significant inter-device communication and leads to underutilization of compute resources, as devices often have to wait for others to finish their part of the computation. To address this, \textit{pipeline parallelism}~\cite{10.5555/3454287.3454297} partitions the model into sequential stages and splits the input data into micro-batches, enabling concurrent execution across devices. Although this improves throughput, it still incurs GPU idle time and demands careful scheduling to maximize resource utilization.

\textit{Tensor parallelism} is a form of model parallelism, which splits individual layers (attention weights across the feature dimension and feedforward layers across the sequence dimension) across devices. While it supports fine-grained parallelism, it requires multiple \texttt{AllReduce} operations per layer, leading to high communication overhead that becomes a bottleneck under limited bandwidth~\cite{Shoeybi2019MegatronLMTM}. To address this, \textit{position-wise partitioning}~\cite{10631032} (a form of data parallelism) has been proposed for Transformer models, where the input sequences/tokens are partitioned across devices instead of the model weights. This reduces communication to a single \texttt{AllGather} per layer, offering improved scalability. Nonetheless, it still suffers from redundant key/value computations across devices and remains sensitive to network bandwidth constraints.

In this work, we propose \name, a communication-efficient and compute-aware extension to position-wise partitioning for distributed inference across edge devices. \name is a Transformer-based model-serving system designed to enable distributed inference across edge networks by efficiently partitioning input/intermediate data and parallelizing computations. Its goal is to minimize communication overhead and redundant computation while maintaining inference accuracy, making it well-suited for deployment in resource-constrained environments. As shown in Fig.~\ref{arch}, our approach introduces the use of a \sm representation to compress the intermediate data exchanged between devices, thereby eliminating the need to transmit full input or intermediate data partitions. Moreover, we restructure the self-attention mechanism to simplify computations and reduce redundant Key/Value matrix operations. For autoregressive models like GPT-2, we introduce a partition-aware causal masking strategy that ensures each device generates the appropriate causal mask for its partition, while preserving communication efficiency.

Our main contributions are summarized as follows.
\begin{itemize}
    \item We introduce the use of \sm representations for compressing input and intermediate feature data, reducing inter-device communication by up to 89.9\% on the ViT model with $P = 2$ and $\textit{CR} = 9.9$. This enables efficient Transformer inference in bandwidth-constrained edge environments, with only a minor impact on accuracy, which can be mitigated through model finetuning.
    \item We eliminate redundant Key and Value matrix computation across devices by limiting it to local partition data and their corresponding received \sm vectors. To preserve the model output shape and accuracy despite the impact of compression on the Key and Value matrices, we restructure the self-attention equation by introducing a scaling-aware softmax computation. This optimization reduces overall computation by up to 50\% and 68\% for 2 and 3 devices, respectively.
    \item We propose a partition-aware causal mask generation strategy based on the self-attention permutation invariance, enabling accurate distributed inference in autoregressive models.
   
\end{itemize}
These contributions offer a scalable and generalizable framework for deploying large Transformer models in distributed resource-constrained edge environments.

The remainder of this paper is organized as follows. Section~\ref{sec:background} provides background on Transformer architectures and review various parallelism techniques relevant to distributed inference. Section~\ref{sec:overview} offers an overview of the \name and its key components. Section~\ref{sec:methodology} presents the detailed methodology, including the \sm representation, self-attention restructuring and partition-aware causal mask. Section~\ref{sec:evaluation} evaluates the \name across ViT, BERT and GPT-2 models, and discusses its limitations and potential future directions. Finally, Section~\ref{conclusion} concludes the article.

Note that we use the terms device and edge device interchangeably throughout the paper.

\noindent \textit{Notations:} Bold uppercase letters (e.g., $\mathbf{X}$) denote matrices, and bold lowercase letters (e.g., $\mathbf{x}$) denote vectors. Italic lowercase letters (e.g., $k$, $n$) represent scalar variables or indices, while italic uppercase letters (e.g., $N$, $D$, $P$) and Greek letters (e.g., $\gamma$, $\sigma$) denote constants or hyperparameters. Calligraphic letters (e.g., $\mathcal{I}$) represent sets. The set of real numbers is represented by $\mathbb{R}$. The operator $\mathbb{S}(\cdot)$ denotes the softmax function, and $\odot$ denotes element-wise multiplication.

\section{Background \& Related Work}\label{sec:background}
In this section, we provide an overview of the Transformer architecture and its components, the challenges associated with large-scale models, and the motivation for distributed inference. We also review key parallelism strategies used to accelerate inference, including model and pipeline parallelism, tensor parallelism, and position-wise layer partitioning.

\subsection{Transformer Model}
The Transformer model was first introduced in \cite{Vaswani2017AttentionIA} for NLP tasks and has since been widely adopted in various domains, including Computer Vision. The architecture is composed of encoder and decoder blocks, each built upon two primary components: the multi-head self-attention mechanism and the feedforward network (FFN). What sets Transformers apart from other computing blocks used in deep learning models is the use of self-attention, which enables the model to capture contextual dependencies between tokens effectively, and is formulated as:
\begin{equation} \label{eq:sda}
\mathbf{A} =\mathbf{S}\mathbf{V}= \mathbb{S} \left( \frac{\mathbf{Q}  \mathbf{K}^T}{\sqrt{d}} \right)  \mathbf{V}, 
\end{equation}

\begin{equation}
\mathbf{Q} = \mathbf{X}\mathbf{W}_q, \:\:\:\: \mathbf{K}=\mathbf{X}\mathbf{W}_k, \:\:\:\: \mathbf{V}=\mathbf{X}\mathbf{W}_v,
\end{equation}

where $ \mathbf{A} \in \mathbb{R}^{N \times d} $ is the output of the self-attention layer, $N$ is the number of input tokens or sequences, and $d$ is the dimensionality of the attention features. The softmax function is denoted by $\mathbb{S}$, and $\mathbf{S} \in \mathbb{R}^{N \times N}$ is the resulting attention score matrix. The input matrix is $\mathbf{X} \in \mathbb{R}^{N \times D}$, where $D$ is the dimensionality of the input embeddings. The learned projection matrices $\mathbf{W}_q, \mathbf{W}_k, \mathbf{W}_v \in \mathbb{R}^{D \times d}$ map the input data to Query, Key, and Value representations, respectively. The projected matrices $\mathbf{Q}, \mathbf{K}, \mathbf{V} \in \mathbb{R}^{N \times d}$ are obtained by multiplying the input $\mathbf{X}$ with these projections.

\begin{figure}
\centering
\includegraphics[width=\columnwidth]{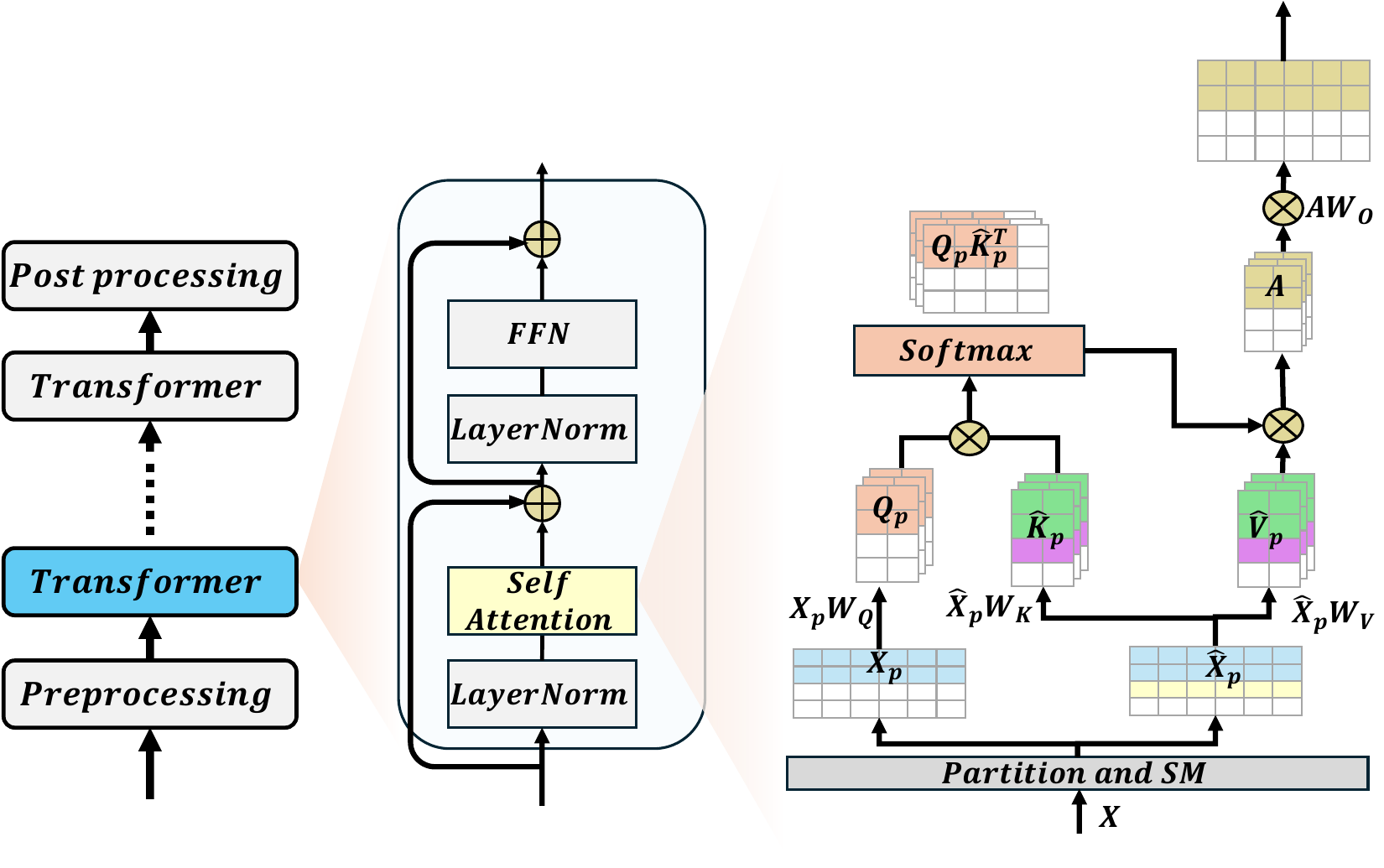}
\caption{General architecture of Transformer-based models (left), Transformer block including the self-attention mechanism (middle), and an example of partitioned self-attention (right).}
\label{transformer}
\end{figure}

Transformer-based FMs can be broadly categorized into two types: encoder-only models, such as BERT~\cite{devlin2019bert} and Vision Transformer (ViT)~\cite{dosovitskiy2021imageworth16x16words}, and decoder-only models, such as GPT-2~\cite{radford2019language,brown2020language}. The main distinction lies in their attention mechanism. Decoder blocks incorporate masked multi-head self-attention, which restricts access to future tokens during both training and inference, enabling autoregressive data generation. The left part of Fig.~\ref{transformer} illustrates the general architecture of Transformer-based models.

Despite their effectiveness, Transformer models are highly memory- and compute-intensive, especially in large-scale configurations with billions of parameters. This makes them difficult to deploy on resource-constrained edge device. 

\subsection{Parallelism Techniques}
The growing scale and complexity of deep learning models, pose significant challenges for resource-constrained environments such as edge device or edge-cloud systems. To address the computational and memory limitations inherent to such platforms, a range of parallelism techniques have been developed. These methods aim to partition the model's workload across multiple processing units, enabling scalable and efficient inference. In the context of edge, cloud, and edge-cloud continuum deployments, parallelism not only facilitates inference acceleration for latency sensitive applications, but also improves the utilization of heterogeneous computing resources. In this subsection, we provide an overview of the main categories of parallelism techniques used for distributed inference, including model parallelism, pipeline parallelism, tensor parallelism, and data parallelism. 



\subsubsection{Model \& Pipeline Parallelism}

Model parallelism~\cite{NIPS2012_6aca9700, 10.5555/3454287.3454297,10096914} divides the model across devices, assigning each device a subset of layers or blocks. After one device completes its computation, the intermediate output is transferred to the next device. To mitigate the communication overhead associated with transferring these intermediate features, prior work in the context of Convolutional Neural Networks (CNNs)\cite{10812766, 10465187} has explored compressing the features before transmission, thereby reducing both latency and data transfer costs. While Model parallelism helps overcome the constraint of deploying large models on a single GPU, it often results in device under utilization, since each device must wait for data from the preceding stage. Pipeline parallelism~\cite{10.5555/3454287.3454297} addresses this inefficiency by splitting input batches into micro-batches, allowing multiple devices to process different micro-batches concurrently. However, pipeline parallelism is effective primarily when batched input data is available. Therefore, for single-query inference, it offers limited benefits due to increased latency and synchronization overhead.

\subsubsection{Tensor Parallelism}
Tensor parallelism~\cite{Shoeybi2019MegatronLMTM} distributes individual layers' parameters across multiple devices by splitting model weights (e.g., within matrix multiplications) and computing them in parallel. According to~\cite{10631032}, this method requires two collective \texttt{AllReduce} operations per Transformer block, leading to per device per layer communication overhead of $4(P - 1)ND/P$, where $ P$ denotes the number of input data partitions (or devices involved in computation). When evaluated on ViT, BERT, and GPT-2 models with $ P = 6 $ participating edge devices, tensor parallelism required a minimum network bandwidth of 1000~Mbps to outperform single-device inference~\cite{10631032}. This method is generally more suitable for distributed inference with smaller batch sizes compared to model or pipeline parallelism, as it enables more effective utilization of device memory and compute resources. Unlike model or pipeline parallelism, where devices may remain idle while waiting for intermediate outputs from others, tensor parallelism allows all devices to compute concurrently on the available input data, thereby minimizing idle time and improving overall resource utilization. However, tensor parallelism requires synchronization across devices twice per Transformer block, i.e., once after the self-attention sub-layer and once after the FFN sub-layer. This inter-device communication imposes a high bandwidth requirement, making it best suited for deployment in cloud environments with fast interconnects like NVIDIA's NVLink whose per link bandwidth ranges from 40~GB/s to 100~GB/s.

\subsubsection{Data Parallelism}

Data parallelism is a foundational paradigm in distributed deep learning, in which the same model is replicated across multiple devices, each processing a distinct subset of the input data in parallel. This approach enables scalable inference by leveraging multiple compute devices to increase throughput. The theoretical foundation of data parallelism can be traced back to the Bulk Synchronous Parallel (BSP) model~\cite{10.1145/79173.79181}, which provides a formal framework for organizing computations into parallel tasks interleaved with synchronized communication phases. Due to its simplicity, scalability, and compatibility with standard hardware accelerators, data parallelism has been widely adopted for large-scale training and inference of Convolutional Neural Networks (CNNs)~\cite{LI2022109150}.

Position-wise layer partitioning~\cite{10631032} for Transformer models is a form of data parallelism, which is motivated by the observation that many Transformer operations, such as FFN and layer normalization are position-wise. In other words, computations on individual tokens or sequence positions can be executed independently. Using this property, the input tensor $\mathbf{X}$ is split along the dimension of the sequence/tokens into partitions $[\mathbf{X}_1;\mathbf{X}_2;...;\mathbf{X}_P]$, which are independently processed across multiple devices. Each device computes the sub-query, key, and value matrices and performs attention and FFN computations locally. This design reduces inter-device synchronization to a single \texttt{AllGather} operation per layer instead of two \texttt{AllReduce} as in tensor parallelism. This reduces communication overhead to $(P-1)ND/P$ per-device per-layer, i.e. reducing thee quarters of communication overhead compared to tensor parallelism in the distributed inference. However, this approach still requires all devices to share their partial outputs after each layer to build complete Key and Value matrices for the next layer's attention computation. This introduces two challenges: (1) redundant Key and Value matrices are generated across devices, and (2) every device must communicate its intermediate outputs to all other participating edge devices. Despite existing optimizations, the communication overhead remains non-trivial. In bandwidth-constrained edge environments, this can lead to significant inference latency and increased communication energy consumption.

To address the challenges of high communication overhead and redundant computation in distributed Transformer inference, we propose \name. \name uses a lightweight compression mechanism using \sm representations that significantly reduces inter-device communication with minor impact on accuracy. It further optimizes computational efficiency by avoiding redundant Key and Value matrix operations and incorporating a scaling-aware attention formulation. Additionally, \name enables accurate inference for autoregressive models through a partition-aware causal masking strategy, making it well-suited for low-latency, bandwidth-constrained edge environments.


\section{System Overview}
\label{sec:overview}

To enable efficient distributed Transformer inference across edge devices, \name introduces a system architecture specifically designed to minimize communication overhead in distributed inference. It follows a master-worker paradigm, where a terminal device (master node) orchestrates data partitioning and coordinates computation across multiple participating edge devices. This design reduces both redundant computation and communication overhead, the two key factors in achieving low-latency inference in distributed settings. Fig.~\ref{arch} illustrates the overall architecture of \name. The system begins with a terminal device (master node), which receives inference requests from users. This master node is responsible for preprocessing the input data and transforming it into an appropriate format for the Transformer model. The master node then partitions the preprocessed data sequence $\mathbf{X}$ into $P$ parts along the sequence dimension $[\mathbf{X}_1;\mathbf{X}_2;...;\mathbf{X}_P]$, where $\mathbf{X}_p \in \mathbb{R}^{N_p \times D}$ and $N_p$ is the number of tokens in $\mathbf{X}_p$ partition. The partitioning procedure is described in Algorithm~\ref{alg:partition}. The number of partitions $P$ corresponds to the number of available edge devices participating in the inference process. After partitioning the input data, the master node also computes \sm representation for each data partition (details provided in Section~\ref{sec:segmented_mean}) and transmits them to the edge devices. In this process, each edge device receives a partition $\mathbf{X}_p$, along with the \sm of the other partitions. For self-attention, each device first computes the Query matrix using its local partition $\mathbf{X}_p$. To compute the Key and Value matrices, the local partition $\mathbf{X}_p$ is combined with the received \sm of rest of the partitions to construct a new matrix $\mathbf{\hat{X}}_p$ as shown in Eq.~\ref{eq:xhat}, and these are then used to compute attention scores as illustrated in the right part of Fig.~\ref{transformer}.

Following the attention operation, each device passes the intermediate data through the FFN to produce the output corresponding to its partition of the Transformer block. As FFN operations are position-wise and independent across tokens, they can be executed without the need of synchronization of data from other devices. When each edge device generates an output of a partition through a Transformer block, it computes its \sm and transmits it to the rest of the edge devices, and likewise it receives \sm from other devices for the next block computation. This process is repeated across all Transformer blocks until the final output is generated.

Compared to the Voltage system \cite{10631032}, which employs position-wise partitioning, \name introduces two significant improvements that further reduce communication and computation overhead:

\begin{itemize}
    \item \textbf{Reduced Communication Overhead:} In existing distributed inference settings, devices must synchronize intermediate outputs after each Transformer block, which introduces latency due to communication delays. \name mitigates this by transmitting only the \sm vectors, substantially reducing communication overhead and improving overall inference latency.
    \item \textbf{Reduced Redundant Computation:} In traditional approaches, each device redundantly computes the full Key and Value matrices using the combined inputs. In contrast, \name limits computation to the local partition and its corresponding \sm, significantly reducing redundant matrix operations and saving compute resources.
\end{itemize}
By addressing these challenges, \name achieves a lower inference latency and more efficient device utilization, making it a practical and scalable solution for Transformer inference across distributed edge environments.

\section{Methodology}\label{sec:methodology}
In this section, we present the methodology behind \name, focusing on how it reduces both communication and computation overhead in Transformer models during distributed inference.
To simplify the description, we concentrate on a single Transformer block and its self-attention mechanism. In Section~\ref{sec:partitioning}, we examine the self-attention operation and its permutation-invariant property, which plays a crucial role in enabling efficient distributed inference. Section~\ref{sec:segmented_mean} describes the calculation of the \sm, a core component in our method for reducing communication overhead. Section~\ref{sec:optimizedSF} discusses the optimization of \sm for self-attention to reduce per device computation. Finally, we address how causal masking is incorporated into \name to support autoregressive models such as GPT-2.
Table \ref{tab:symbols} summarizes the notation used throughout this paper.

\begin{table}[htbp]
\caption{Key Notations and Their Definitions. \label{tab:symbols}}
\centering
\begin{tabular}{ll}
\toprule
\textbf{Notation} & \textbf{Description} \\
\midrule
$N$ & Number of tokens in the input sequence \\
$D$ & Features per token/ Embedding \\
$\mathbf{X}$ & Input matrix of dimensions $N \times D$ \\
$\mathbb{S}$ & Softmax function \\
$\mathbf{X}_p$ & The partitioned segments of $X$ \\
$s$ & Segment size calculated as $\lfloor \frac{N_p}{l} \rfloor$ \\
$\mathbf{Z}_p$ & \sm calculations for each partition $\mathbf{X}_p$ \\
$\mathbf{W}_q, \mathbf{W}_k$ & Weight matrices for query and key transformation \\
$\mathbf{U}$ & $\mathbf{W}_q \mathbf{W}_k^T / \sqrt{D}$ \\
$L$ & Number of \sm per partition\\
$\textit{CR}$ & Compression Rate\\
$P$ & Total number of partitions/devices\\
$p$ & Edge device ID\\
$N_p$ & Number of tokens in $\mathbf{X}_p$ partition \\
$\mathbf{\hat{X}}_p$ & Modified version of $\mathbf{X}$ for device $p$ \\
$\hat{N}_p$ & Number of tokens in $\mathbf{\hat{X}}_p$ \\
$\mathbf{P}_m$ & Permutation Matrix \\
$\mathbf{x}_{p,j}, \mathbf{\hat{x}}_{p,j}, \mathbf{z}_{p,j}$ & Row vectors and $p$ represents the partition ID and\\
    & $j$ represents the row of that partition\\
$\alpha, \lambda$ & Number of times $z_{x,y}$ were repeated\\
$\mathbf{S}_{p}$ & Softmax result of the matrix multiplication \\
    & $\mathbf{X}_p \mathbf{U} \mathbf{\hat{X}}_p^T$ \\
\bottomrule
\end{tabular}
\end{table}

\subsection{Self-Attention and Permutation-Invariance} \label{sec:partitioning}

To enable distributed inference across edge devices, we adopt \textit{position-wise layer partitioning}. However, \name extends this technique further by addressing the communication challenges in standard position-wise partitioning. Specifically, we propose the use of an optimized mechanism that systematically compresses communication data. 

Let us first expand the self-attention equation to better explain the dependencies upon input data:
\begin{equation} 
    \mathbf{S} = \mathbb{S}\left( \hspace{-1mm}\frac{1}{\sqrt{d}} \hspace{-1mm}\left( \hspace{-1mm} \begin{bmatrix}
    \mathbf{x}_1\mathbf{W}_q  \mathbf{W}_k^T\mathbf{x}_1^T \hspace{-1mm}& \mathbf{x}_1\mathbf{W}_q  \mathbf{W}_k^T\mathbf{x}_2^T \hspace{-1mm}& \cdots 
    \\
    \mathbf{x}_2\mathbf{W}_q  \mathbf{W}_k^T\mathbf{x}_1^T \hspace{-1mm}& \mathbf{x}_2\mathbf{W}_q  \mathbf{W}_k^T\mathbf{x}_2^T \hspace{-1mm}& \cdots 
    \\
    \cdots \hspace{-1mm}& \cdots \hspace{-1mm} & \cdots 
    \\
    \mathbf{x}_N\mathbf{W}_q  \mathbf{W}_k^T\mathbf{x}_1^T \hspace{-1mm}& \mathbf{x}_N\mathbf{W}_q  \mathbf{W}_k^T\mathbf{x}_2^T \hspace{-1mm}& \cdots 
    \\
    \end{bmatrix} \hspace{-1mm}\right) \hspace{-1mm}\right) ~.
\end{equation}
For the first row in $\mathbf{Q}\mathbf{K}^T$:


\begin{eqnarray} 
    \mathbf{S}_1 &=& \mathbb{S} \left( \frac{1} {\sqrt{d}}
    \begin{bmatrix}
    \mathbf{x}_1\mathbf{W}_q  \mathbf{W}_k^T\mathbf{x}_1^T &  \cdots & \mathbf{x}_1\mathbf{W}_q  \mathbf{W}_k^T\mathbf{x}_N^T\\
    \end{bmatrix} \right) \label{eq:sdadevice1} \nonumber\\
    &=& \mathbb{S} \left( \frac{1} {\sqrt{d}}\left( \begin{bmatrix}
    \mathbf{x}_1\mathbf{W}_q  \mathbf{W}_k^T\begin{bmatrix}
    \mathbf{x}_1 \\
    \mathbf{x}_2\\
    \cdots\\
    \mathbf{x}_N\\
    \end{bmatrix}^T\\
    \end{bmatrix}\right) \right), \label{eq:expansion}
\end{eqnarray}
where $\mathbf{x}_n$ is the $n^{th}$ row of input matrix $\mathbf{X}$. Eq. \ref{eq:expansion} shows that, for each row in $\mathbf{Q}\mathbf{K}^T$ matrix, we need whole input data $\mathbf{X}$ for the Key matrix, this means that if we follow the standard position-wise partitioning technique, it would require the synchronization of output of partitioned data after every Transformer block.

\begin{figure*}[htbp]
\centering
\subfloat[]{\includegraphics[width=3in,height=1.7in]{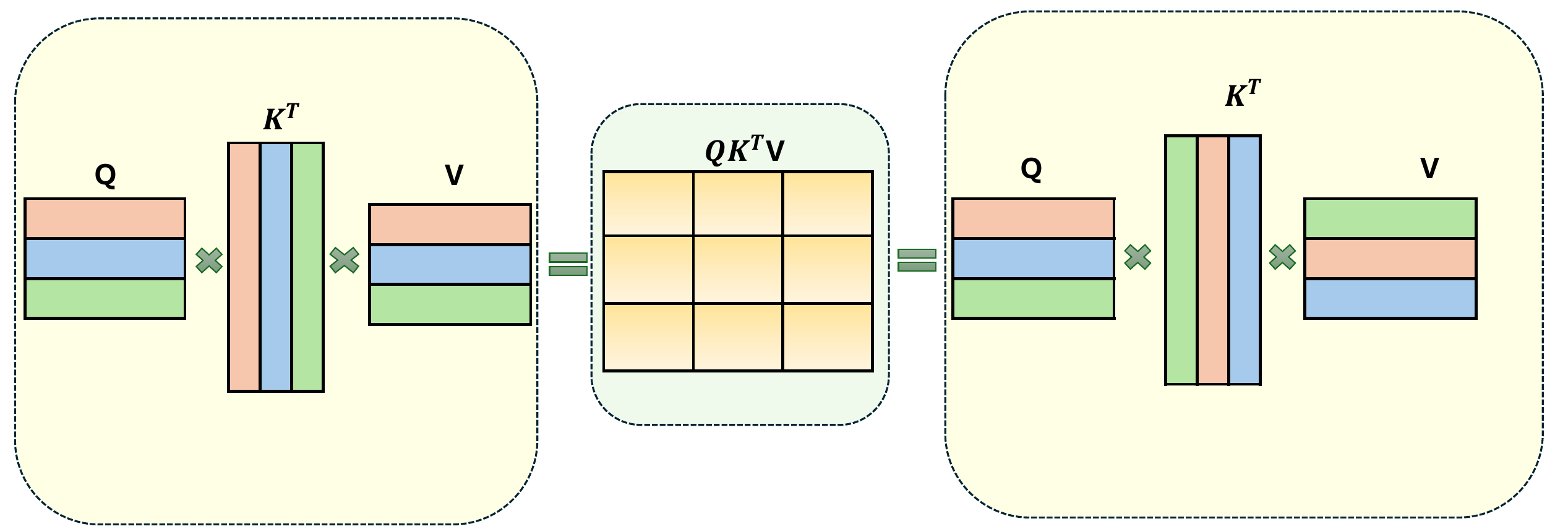}%
\label{fig:permutation_invariance}}
\hfil
\subfloat[]{\includegraphics[width=2in,height=1.7in]{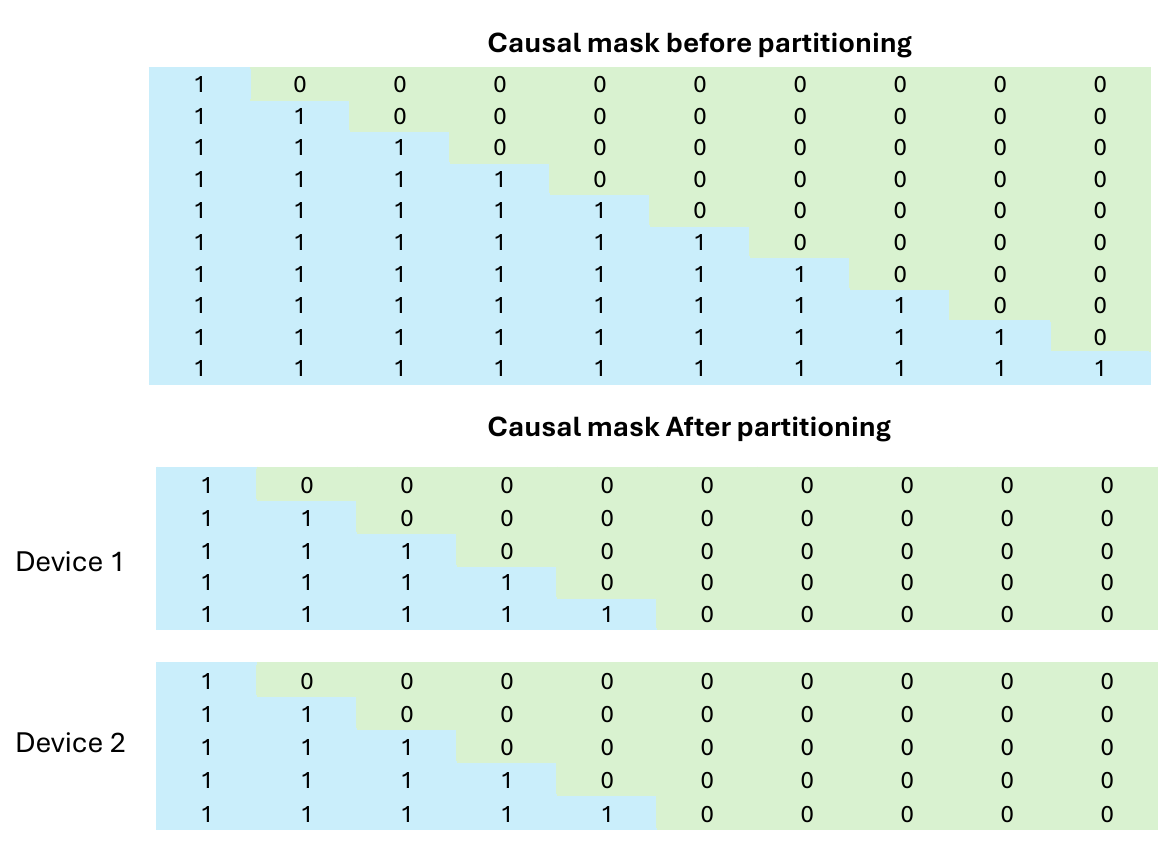}%
\label{fig:issuecausl_mask}}
\hfil
\subfloat[]{\includegraphics[width=2in,height=1.7in]{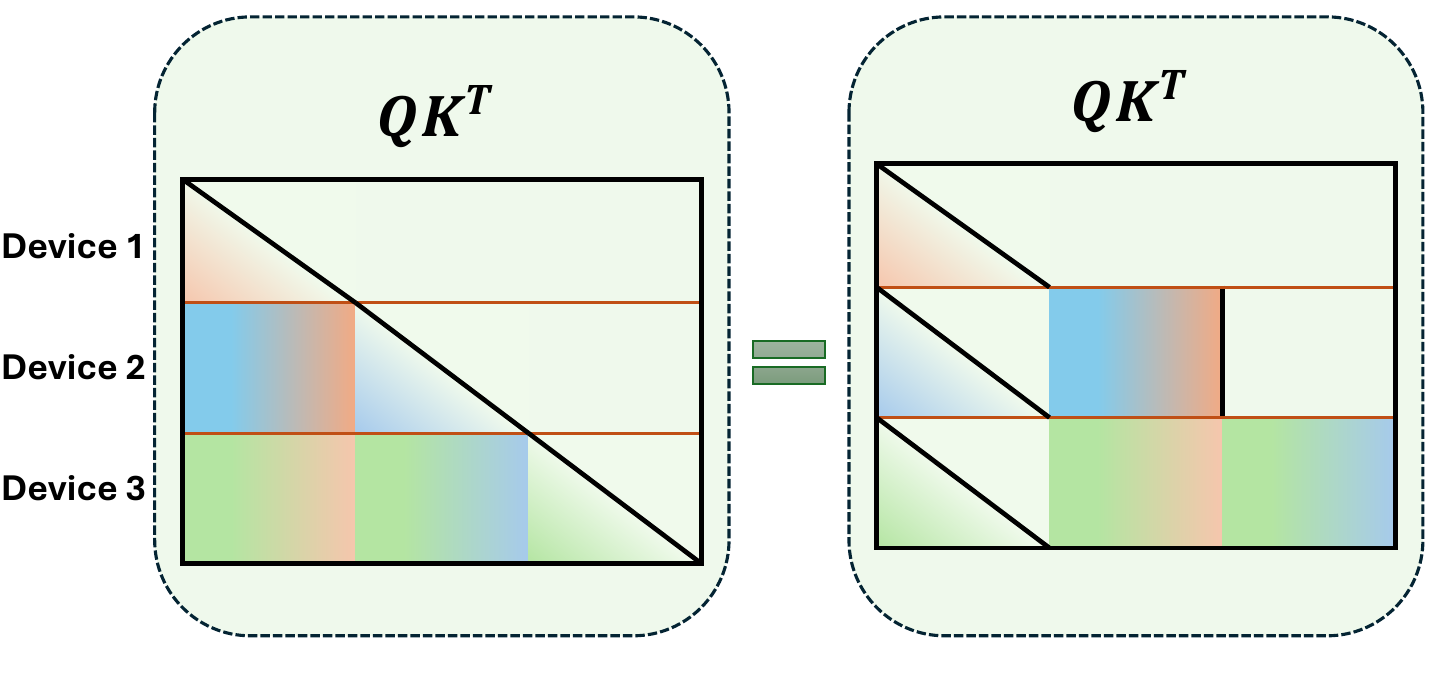}%
\label{fig:causl_mask}}
\caption{(a) The attention mechanism is invariant to row-wise permutations of the Key and Value matrices, enabling flexible and distributed computation. (b) In autoregressive models such as GPT-2, a causal mask is required to prevent attention to future tokens. In a distributed setup, if not handled properly, this can lead to incorrect masking of past and present tokens. (c) We adapt the causal mask for each partition, leveraging the permutation-invariance property, to ensure that only future tokens are masked appropriately on each device.}
\label{fig_sim}
\end{figure*}

Furthermore, it is worth noting that (as illustrated in Fig.~\ref{fig:permutation_invariance}) the self-attention mechanism is inherently permutation invariant with respect to the Key and Value matrices, due to the associativity of matrix multiplication and its invariance under conjugation by permutation matrices:
\begin{equation}\label{eq:permut}
\mathbf{A}=\mathbb{S}\left(\frac{\mathbf{Q}(\mathbf{K}^T \mathbf{P}_m)}{\sqrt{d}}\right)(\mathbf{P}_m^{-1}\mathbf{V})=\mathbb{S}\left(\frac{\mathbf{Q}  \mathbf{K}^T}{\sqrt{d}}\right)\mathbf{V}~,
\end{equation}
where $\mathbf{P}_m$ is an orthogonal permutation matrix. This invariance allows each device to operate on a reordered view of the input data without affecting the final result, forming the foundation of our distributed attention computation. When devices share their respective \sm (detailed in Section~\ref{sec:segmented_mean}), the order of those representations to form $\mathbf{\hat{X}}_p$ is irrelevant due to this property.

To perform attention, each edge device $p$ constructs an augmented matrix $\mathbf{\hat{X}}_p \in \mathbb{R}^{\hat{N}_p\times D}$ having $\hat{N}_p$ rows by concatenating its local partition data $\mathbf{X}_p \in \mathbb{R}^{N_p\times D}$ with the \sm received from the other edge devices, i.e., $\mathbf{Z}_{\hat{p}} \in \mathbb{R}^{L\times D}$, $\hat{p} \in \mathcal{I} \setminus \{p\}$, where $\mathcal{I} = \{1, 2, \dots, P\}$:
\begin{align}\label{eq:xhat}
\mathbf{\hat{X}}_p  &= concat(\mathbf{X}_p, \mathbf{Z}_j, \mathbf{Z}_k, \dots), \notag \\
&\quad \forall p \in \mathcal{I}, \:\:\:\:
(j, k,\dots) \in \mathcal{I} \setminus \{p\} ~.
\end{align}
Each device then computes its Key and Value matrices using its own augmented input $\mathbf{\hat{X}}_p$, while the Query is computed using its local partition data $\mathbf{X}_p$. This selective augmentation reduces redundant computation and enables scalable, parallel self-attention across devices. Based on this, Eq.~\ref{eq:permut} can be written as

\begin{equation} 
    \mathbf{A} = \mathbb{S} \left( \frac{\begin{bmatrix}
    \mathbf{X}_p\mathbf{W}_q  \mathbf{W}_k^T  \mathbf{\hat{X}}_p^T \\
    \end{bmatrix}}{\sqrt{d}} \right) \mathbf{\hat{X}}_p  \mathbf{W}_v = \mathbf{S}_p\mathbf{\hat{X}}_p\mathbf{W}_v ~.
\end{equation}
To simply the equation representation in the next subsection, we use $\mathbf{U}=\mathbf{W}_q\mathbf{W}_k^T/\sqrt{d}$.

\subsection{Segment Means}
\label{sec:segmented_mean}


We adopt \sm-based representations as mechanism to compress intermediate features and reduce inter-device communication during distributed inference. By summarizing the token sequences, \sm enable devices to exchange compact contextual information with minimal overhead. 

The procedure begins by dividing each partitioned input sequence into non-overlapping segments of approximately equal size. For a given partition $\mathbf{X}_p \in \mathbb{R}^{N_p \times D}$, its output is compressed to $L$ vectors. Let $r = (N_p \bmod L)$ denote the remainder, and $s = \lfloor N_p / L \rfloor$ be the base segment size.
Therefore, each segment $l \in \{0, 1, \dots, L - 1\}$ is defined as:


\begin{equation}
\mathbf{H}_l = 
\begin{cases}
\mathbf{X}_p{[ls : (l+1)s]}, & \text{if } l < L - 1  \in \mathbb{R}^{s\times D},\\
\mathbf{X}_p{[ls : (l+1)s + r]}, & \text{if } l = L - 1 \in \mathbb{R}^{(s+r)\times D}.
\end{cases}
\end{equation}
For each segment $\mathbf{H}_l $, the column-wise mean $\mu_l \in \mathbb{R}^{1\times D}$ is computed. This yields the compressed representation:
\begin{equation} \label{eq:nrSM}
\mathbf{Z}_p = [\mathbf{\mu}_0; \mathbf{\mu}_1; \dots; \mathbf{\mu}_{L - 1}] \in \mathbb{R}^{L \times D}. 
\end{equation}

Each $\mathbf{\mu}_l$ serves as a compact representation of the local information within its corresponding segment. The \sm computation procedure is outlined in Algorithm~\ref{alg:segmean}. For the first Transformer block \sm are computed by the master node and shared among devices.

\begin{algorithm}[t]
\caption{Partitioning Input Data Along Sequence Dimension}
\label{alg:partition}
\textbf{Input:} Sequence tensor $\mathbf{X} \in \mathbb{R}^{B \times N \times D}$, number of partitions $P$ \\
\textbf{Output:} List of $P$ tensor partitions $\{\mathbf{X}_1; \mathbf{X}_2; \ldots; \mathbf{X}_P\}$

\begin{algorithmic}[1]
\State $B, N, D \gets$ shape of $\mathbf{X}$
\State $s \gets \lfloor N / P \rfloor$ \Comment{ Base partition size} 
\State $r \gets N \mod P$ \Comment{Remaining elements} 
\State Initialize an empty list $\texttt{partitions}$
\State $start \gets 0$ \Comment{Start index}
\For{$i = $1 to $P$}
    \State $end \gets start + s$
    \If{$i = P $}
        \State $end \gets end + r$ \Comment{Last partition takes remainder}
    \EndIf
    \State $\mathbf{X}_i \gets \mathbf{X}[:, start:end, :]$ 
    \State Append $\mathbf{X}_i$ to $\texttt{partitions}$
    \State $start \gets end$
\EndFor
\State \textbf{Return} $\texttt{partitions}$
\end{algorithmic}
\end{algorithm}

\begin{algorithm}[t]
\caption{\sm Computation for Sequence Compression}
\label{alg:segmean}
\textbf{Input:} Partitioned Sequence tensor $\mathbf{X}_p \in \mathbb{R}^{B \times N_p \times D}$, number of landmarks $L$ \\
\textbf{Output:} Compressed tensor $\mathbf{Z}_{\text{p}} \in \mathbb{R}^{B \times L \times D}$
\begin{algorithmic}[1]

\State $B, N_p, D \gets$ shape of $\mathbf{X}$
\State $\text{r} \gets N_p \bmod L$
\State $\text{s} \gets \lfloor N_p / L \rfloor$
\State Initialize empty list $\text{segments}$ 
\For{$i \gets 0$ to $L-1$}
    \If{$i < L - 1\ \mathbf{or} \  \text{r} = 0$}
        \State $start \gets i \cdot \text{s}$
        \State $end \gets start + \text{s}$
    \Else
        \State $start \gets i \cdot \text{s}$
        \State $end \gets start + \text{s} + \text{r}$
    \EndIf
    \State $\mathbf{H}_l \gets \mathbf{X}_p[:, start:end, :]$ \Comment{Extract segment}
    \State $m \gets \text{Mean}(S, \text{dim}=-2)$ \Comment{Mean over sequence length}
    \State Append $m$ to $\text{segments}$
\EndFor
\State $Z_{\text{p}} \gets \text{stack}(\text{segments}, \text{dim}=1)$
\State \textbf{Return} $Z_{\text{p}}$ 
\end{algorithmic}
\end{algorithm}

\subsection{Attention Optimization}
\label{sec:optimizedSF}

Once \sm are computed, each edge device exchanges its local \sm $\mathbf{Z}_p$ with the other edge devices involved in the distributed inference. Upon receiving \sm from all other edge devices, each edge device constructs $\mathbf{\hat{X}}_p$ as defined in Eq.~\ref{eq:xhat}, and uses it for the calculation of the Key and Value matrices of next layer's self-attention computation: 
\begin{equation} \label{eq:singlerowattention}
    \mathbf{S}_{p,1} = \frac{1}{\sum_{j=1}^{\hat{N}_p} e^{\mathbf{x}_{p,1} \mathbf{U} \mathbf{\hat{x}}_{p,j}^T}} \left( e^{\mathbf{x}_{p,1} \mathbf{U} \mathbf{\hat{x}}_{p,1}^T}, \dots, e^{\mathbf{x}_{p,1} \mathbf{U} \mathbf{\hat{x}}_{p,\hat{N}_p}^T} \right)~,
\end{equation}
\noindent where $\mathbf{S}_{p,1}$ denotes the first row of the softmax matrix for the $p$-th partition. The reduced size of the Key and Value matrices can impact the overall model accuracy. To better approximate the original structure of these matrices, each device duplicates every vector in $\mathbf{Z}_p$ according to the number of tokens in its corresponding original segment after receiving the \sm representations from the other devices. This duplication strategy helps improve model accuracy, as demonstrated in Table~\ref{tab:duplicates}. The expanded representation becomes:
\begin{equation} \label{eq:rSM}
\mathbf{Y}_p = [\mathbf{\mu}_0^{(1)}; \dots; \mathbf{\mu}_0^{(n_0)}; \mathbf{\mu}_1^{(1)}; \dots; \mathbf{\mu}_1^{(n_1)}; \dots; \mathbf{\mu}_{L - 1}^{(n_{L-1})}]~,
\end{equation}
where $\mathbf{\mu}_l^{(i)} = \mathbf{\mu}_l$ for $i = 1, \dots, n_l$, and $\mathbf{Y}_p \in \mathbb{R}^{N_p \times D}$.

\begin{table}[t]
\caption{Impact of Duplicated \sm Vectors on ViT Self-Attention (CIFAR-10). \label{tab:duplicates}}
\centering
\begin{tabular}{ccccc}
\toprule
\multirow{3}{*}{\textbf{P}} & \multirow{2}{*}{\textbf{PDPLC}} &\multirow{3}{*}{\textbf{CR}}& \multicolumn{2}{c}{\textbf{Duplicated?}} \\
\cline{4-5}
&&&\textbf{No}&\textbf{Yes}\\
\cline{4-5}
&\textbf{Tokens}&&\textbf{Acc.}&\textbf{Acc.}\\

\midrule
2&10&9.90&91.66&\textcolor{blue}{\textbf{95.64}}\\
2&20&4.95&95.4&\textcolor{blue}{\textbf{96.84}}\\
2&30&3.30&96.48&\textcolor{blue}{\textbf{97.06}}\\
\bottomrule
\end{tabular}
\end{table}

Compared to Voltage \cite{10631032}, which requires each device to exchange $\left(P - 1\right)ND/P$ elements per layer, \name reduces this to $\left(P - 1\right) L D$, where $L \ll N/P$ and depends on the compression rate. While substantially reducing communication cost, \name allows an efficient approximation of the input features during the generation of Key and Value matrices, resulting to a practical balance between accuracy and inference time.

Until this step, the computational workload per device remains comparable to that in Voltage \cite{10631032}, because the Key and Value matrices retain their original dimensions (i.e., $\mathbf{K}, \mathbf{V} \in \mathbb{R}^{N \times d}$). To address this, we further optimize computation by leveraging the repetitive structure of the expanded matrix $\mathbf{Y}_p$. Upon closer analysis, we observed that the repetitive structure of vectors in \( \mathbf{Y}_p \) from Eq.~\ref{eq:rSM} introduces redundancy within the attention computation. By leveraging this duplication of vectors, the self-attention equation Eq.~\ref{eq:singlerowattention} can be restructured to eliminate redundant operations, thereby reducing computational overhead without affecting the output or the accuracy. Specifically, we reformulate the attention mechanism by factoring out repeated components and reusing intermediate computations. Rather than recomputing identical exponentiated terms for each duplicated vector, we introduce scaling factors into the softmax expression to efficiently account for the repeated segments. This optimization avoids redundant operations while preserving the dimensions of the attention mechanism. 

Substituting $\mathbf{Y}_p$ from Eq.~\ref{eq:rSM} into Eq.~\ref{eq:singlerowattention}, and letting $\phi = e^{\mathbf{x}_{p,1} {U}}$, we leverage the property of Exponentiation Associativity $x^{ab} = (x^a)^b$ to simplify repeated operations and eliminate redundant computation as follows:

\begin{equation} \label{eq:simplifiedsinglerowattention}
\begin{aligned}
\mathbf{S}_{p,1} = \frac{
\left( \phi^{\mathbf{\hat{x}}_{1,1}^T}, \dots, \phi^{\mathbf{\hat{x}}_{1,\hat{N}_p}^T} \right)
}{
\displaystyle\sum_{k=1}^{N_p} \phi^{\mathbf{x}_{p,k}^T} 
+ \displaystyle\sum_{k=1}^{L-1} \alpha \sum_{\substack{j=1 \\ j \ne p}}^P \phi^{\mathbf{y}_{j,k}^T}
+ \displaystyle\sum_{\substack{j=1 \\ j \ne p}}^P \lambda_j \phi^{\mathbf{y}_{j,L}^T}
} ~.
\end{aligned}
\end{equation}

From Eq.~\ref{eq:simplifiedsinglerowattention}, two important observations emerge: (1) the denominator includes scaling factors ($\alpha, \lambda \geq 1$) to account for duplicated vectors; and (2) redundant elements still exist in both the numerator and the Value matrix. To address this, we use a non-duplicated \sm matrix as defined in Eq.~\ref{eq:nrSM}, along with a repetition count (scaling) vector $\mathbf{g} \in \mathbb{R}^{1 \times N}$ that records how many times each row in $\mathbf{\hat{X}}$ should be repeated, instead of actually duplicating it, which avoids redundant computation. We start it by computing the exponential matrix once and scaling it using $\mathbf{g}$:

\begin{equation}
    \mathbf{\Psi} = \exp \left( \frac{\mathbf{X}_p \mathbf{W}_q \cdot \mathbf{W}_k^T \mathbf{\hat{X}}_p^T}{\sqrt{d}} \right)=\exp \left( \frac{\mathbf{Q}_p \mathbf{\hat{K}}_p^T}{\sqrt{d}} \right),
\end{equation}

\begin{equation}
    \mathbf{\mathcal{E}} = [\mathbf{\Psi}\odot \mathbf{g}]_{i,j} = \mathbf{\Psi}_{i,j} \cdot \mathbf{g}_j,
\end{equation}

\begin{equation}\label{eq:simplifiedattention}
    \mathbf{A} = \mathbb{S}(\mathbf{\mathcal{E}}) \cdot \mathbf{\hat{X}}_p \mathbf{W}_v=\mathbb{S}(\mathbf{\mathcal{E}}) \cdot \mathbf{\hat{V}}_p.
\end{equation}

Here, \( \odot \) denotes element-wise (Hadamard) multiplication,  and this broadcasts \( \mathbf{g} \) across all rows of \( \Psi\). 
It is worth mentioning that, in the case of large language models (LLMs), where input sequences vary in length (unlike in ViT, which often uses fixed-size inputs), it is appropriate to compute the number of non-repeated \sm  vectors per partition $L$ dynamically, based on the input sequence length and compression rate, i.e.:
\begin{equation} 
    L = \left\lfloor \frac{N}{\textit{CR} \cdot P}\right\rfloor~.
\end{equation}

$\textit{CR}$ is compression rate used to control the degree of compression applied to the intermediate output features. It determines the trade-off between reduced communication data/computation and accuracy loss. The formulation represented in Eq.~\ref{eq:simplifiedattention} not only avoids redundant exponentiations but also reduces the required number of matrix multiplications. By exploiting the repetitive structure of \sm, we achieve reduced computation cost while preserving the correctness of the attention mechanism.

\subsection{Causal Mask}
\label{sec:causal_mask}

Autoregressive language models, such as GPT-2, apply a \textit{causal mask} to ensure that each token only attends to itself and its preceding tokens, thereby preserving the left-to-right generation constraint. In a distributed inference setup, where the input sequence is partitioned across multiple devices, causal masking must be carefully handled to prevent incorrect attention behavior. 
As shown in Fig.~\ref{fig:issuecausl_mask}, naively applying standard lower-triangular masking on each device's local partition data leads to incorrect masking. Since each device operates on a subset of the full sequence, applying a standalone mask causes it to erroneously mask valid positions, including past and current tokens (see Fig.~\ref{fig:issuecausl_mask}). 
To avoid this, each device must construct its causal mask using its global position within the original sequence of length $N$. The master device communicates the global partition index $p$ of each partition $[\mathbf{X}_1; \mathbf{X}_2; \ldots; \mathbf{X}_P]$ to the respective devices. This enables each device to build a mask that only prevents attention to future tokens across all partitions.

Let $\mathbf{M}_p \in \mathbb{R}^{N_p \times \hat{N}_p}$ be the mask applied on $\mathbf{Q}\mathbf{K}^T$ on device \( p \). We define the causal mask $\mathbf{M}_p$ as:

\begin{equation}
\mathbf{M}_p[i,j] = 
\begin{cases}
1, & \text{if } 0 \leq j \leq i < N_p, \\
1, & \text{if } N_p \leq j < N_p + L \cdot (p - 1), \\
0, & \text{otherwise}.
\end{cases} 
\end{equation}

This ensures that each row in the local partition's attention matrix only allows attention to its own and all preceding tokens across the global sequence, regardless of the out-of-order reception of segments. As illustrated in Fig.~\ref{fig:causl_mask}, the adjusted masks correctly account for the position of each partition, enabling devices to preserve causal structure in a distributed and potentially out-of-order processing environment.

\section{Experiment and Performance Evaluation} 
\label{sec:evaluation}





\subsection{Experimental Setup}
\label{sec:experimental_setup}

To evaluate the effectiveness of the proposed distributed inference strategy, we conducted experiments on NVIDIA RTX 2080, RTX 2080 Ti, and RTX A6000 GPUs to simulate a distributed environment. We tested the scenarios of distributed inference with two and three devices, i.e., $P = 2$ and $P = 3$, where each device independently processes a subset of the input data and exchanges the intermediate features at the end of each Transformer block to collaboratively perform the inference.
We evaluated \name on three widely-used Transformer architectures, namely Vision Transformer (ViT)~\cite{dosovitskiy2021imageworth16x16words}, BERT~\cite{devlin2019bert}, and GPT-2~\cite{radford2019language, brown2020language}. The evaluation setup was tailored to each model and its associated tasks and datasets, as detailed below.

\subsection{Datasets}
\label{sec:datasets}

We evaluated \name using a diverse collection of datasets suited for vision and language tasks as shown in Table~\ref{tab:datasets}. For image classification with Vision Transformer (ViT), we used CIFAR-10, CIFAR-100~\cite{krizhevsky2009learning}, and ImageNet-1K~\cite{5206848}. CIFAR-10 and CIFAR-100 each contain 50,000 training and 10,000 test color images of size \(32 \times 32\), with 10 and 100 classes, respectively. ImageNet-1K is a large-scale dataset containing approximately 1.2 million training images and 50,000 validation images across 1,000 categories.

For natural language understanding (NLU) tasks, we utilized the General Language Understanding Evaluation (GLUE) benchmark~\cite{wang-etal-2018-glue} suite to evaluate BERT. This includes Multi-Genre Natural Language Inference (MNLI), Question answering Natural Language Inference (QNLI), and Recognizing Textual Entailment (RTE), which are natural language inference tasks requiring the classification of sentence pairs into entailment, contradiction, or neutral. Microsoft Research Paraphrase Corpus (MRPC) and Quora Question Pairs (QQP) are paraphrase detection tasks where the objective is to identify whether two sentences are semantically equivalent. Stanford Sentiment Treebank (SST-2) is used for binary sentiment classification, while Corpus of Linguistic Acceptability (CoLA) focuses on linguistic acceptability. Semantic Textual Similarity Benchmark (STS-B) provides semantic similarity scores between sentence pairs on a scale from 0 to 5.

For language modeling using GPT-2, we used Children’s Book Test-Common Noun (CBT-CN) and Children’s Book Test-Named Entities (CBT-NE)~\cite{hill2015goldilocks}, which are cloze-style datasets targeting common nouns and named entities, respectively. Additionally, we employed two character-level modeling datasets to assess compression: enwik8, which consists of 100 MB of unprocessed Wikipedia text, evaluated using bits per byte (bpb), and text8, a preprocessed version of Wikipedia text containing 100 MB of lowercase characters only, evaluated using bits per character (bpc).



\begin{table}[!htbp]
\caption{Summary of datasets, evaluation metrics, and models used in experiments. \label{tab:datasets}}
\centering
\begin{tabular}{lll}
\toprule
\textbf{Dataset} & \textbf{Evaluation Metric} & \textbf{Model} \\
\midrule
CIFAR-10 & Accuracy & ViT \\
CIFAR-100 & Accuracy & ViT \\
ImageNet-1K & Top-1 Accuracy & ViT \\
\midrule
MNLI & Accuracy & BERT \\
QNLI & Accuracy & BERT \\
RTE & Accuracy & BERT \\
MRPC & F1 Score & BERT \\
QQP & F1 Score & BERT \\
SST-2 & Accuracy & BERT \\
CoLA & MCC & BERT \\
STS-B & Spearman Correlation & BERT \\
\midrule
CBT-CN & Accuracy & GPT-2 \\
CBT-NE & Accuracy & GPT-2 \\
enwik8 & BPC & GPT-2 \\
text8 & BPC & GPT-2 \\
\bottomrule
\end{tabular}
\end{table}

\subsection{Evaluation Metrics}
\label{sec:metrics}

We employed a variety of task-specific metrics to evaluate model performance under distributed inference configurations. For the ImageNet-1K, CIFAR-10, CIFAR-100, MNLI, QNLI, RTE, SST-2, CBT-NE, and CBT-CN classification tasks, we used accuracy defined as the ratio of correctly predicted samples to the total number of samples:
\begin{equation}
\text{Accuracy} = \frac{\text{TP} + \text{TN}}{\text{TP} + \text{TN} + \text{FP} + \text{FN}},
\end{equation}
where TP, TN, FP, and FN represent true positives, true negatives, false positives, and false negatives, respectively.

For the MRPC and QQP tasks which have class imbalance, we used the F1 score. It is the harmonic mean of precision and recall, where precision and recall are computed as follows:
\begin{equation}
\text{Precision} = \frac{\text{TP}}{\text{TP} + \text{FP}}, \quad \quad
\text{Recall} = \frac{\text{TP}}{\text{TP} + \text{FN}},
\end{equation}
\begin{equation}
\quad
\text{F1} = 2 \cdot \frac{\text{Precision} \cdot \text{Recall}}{\text{Precision} + \text{Recall}} ~.
\end{equation} 

For CoLA, a task involving linguistic acceptability, we used the Matthews Correlation Coefficient (MCC)~\cite{matthews1975comparison}, which provides a balanced measure even in the presence of imbalanced classes:
\begin{equation}
\text{MCC} = \frac{(\text{TP} \cdot \text{TN}) - (\text{FP} \cdot \text{FN})}{\sqrt{(\text{TP}+\text{FP})(\text{TP}+\text{FN})(\text{TN}+\text{FP})(\text{TN}+\text{FN})}}.
\end{equation}

To evaluate semantic similarity in the STS-B dataset, we applied the Spearman rank correlation coefficient~\cite{spearman1961proof}, which assesses the monotonic relationship between predicted and human-annotated similarity scores:
\begin{equation}
\rho = 1 - \frac{6 \sum d_i^2}{n(n^2 - 1)}~,
\end{equation}
where \( d_i \) is the difference in ranks and \( n \) is the number of paired samples.

For evaluating character-level language modeling and compression efficiency, we used bits per byte (BPB) for enwik8 and bits per character (BPC) for text8 datasets~\cite{blevins-zettlemoyer-2019-better}. These are computed as:

\begin{equation}
\text{BPB} = -\frac{1}{h_{byte}} \sum_{i=1}^{h_{byte}} \log_2 \eta(b_i)~,
\end{equation}
\begin{equation}
\text{BPC} = -\frac{1}{h_{char}} \sum_{i=1}^{h_{char}} \log_2 \eta(c_i)~,
\end{equation}

\noindent where $h_{byte}$ and $h_{char}$ are the total numbers of bytes and characters respectively in a sequence, $\eta(b_i)$ and $\eta(c_i)$ are the model's cross-entropy loss for byte $b_i$ and character $c_i$ respectively. We also examined various compression rates ranging from $\textit{CR}=2$ to $\textit{CR}=10$ across different partition settings. This allowed us to analyze the trade-offs between communication efficiency, computation speed-up, and the preservation of task-specific accuracy in distributed inference.

\begin{table*}[!ht]
\centering
\caption{Comparison of Computation and Communication Efficiency For ViT Model}
\label{tab:strategy_comparison}
\begin{tabular}{lcm{1.2cm}m{1.2cm}m{1.2cm}m{1.2cm}m{1.2cm}m{1.2cm}m{1.2cm}m{1.2cm}m{1.2cm}}
\toprule
\multirow{2}{*}{\textbf{Strategy}} & \multirow{2}{*}{\textbf{P}} & \textbf{GFLOPs} & \textbf{GFLOPs} & \textbf{Comp. Speed-up} & \textbf{PDPLC} & \multirow{2}{*}{\textbf{CR}} & \textbf{Comm. Speed-up} & \textbf{CIFAR-10} & \textbf{CIFAR-100} & \textbf{ImageNet-1K} \\
\cline{3-6}\cline{8-11}
&&\textbf{Total}& \textbf{/device}&\textbf{\%}&\textbf{Tokens}&&\textbf{\%}&Acc.&Acc.&Acc.\\
\midrule
No partition & 1 & 35.15 & 35.15 & - & - & - & - & 98.01 & 91.00 & 80.30 \\
Voltage \cite{10631032} & 2 & 40.74 & 20.37 & 42.05 & 99 & - & - & 98.01 & 91.00 & 80.30 \\
Voltage \cite{10631032} & 3 & 46.33 & 15.44 & 56.06 & 131 & - & - & 98.01 & 91.00 & 80.30 \\
\midrule
\multirow{6}{*}{\textbf{\name}} 
& 2 & 35.07 & 17.54 & 50.11 & 10 & 9.90 & 89.90 & 95.64 & 85.25 & 72.64 \\
& 2 & 35.71 & 17.86 & 49.20 & 20 & 4.95 & 79.80 & 96.84 & 88.00 & 76.45 \\
& 2 & 36.35 & 18.18 & 48.29 & 30 & 3.30 & 69.70 & 97.06 & 89.20 & 77.68 \\
& 3 & 36.04 & 12.01 & 65.82 & 20 & 6.55 & 84.73 & 94.49 & 84.07 & 70.70 \\
& 3 & 37.89 & 12.63 & 64.07 & 40 & 3.28 & 69.47 & 96.60 & 87.80 & 76.35 \\
& 3 & 39.73 & 13.24 & 62.32 & 60 & 2.18 & 54.20 & 97.03 & 88.94 & 77.98 \\
\midrule 
\multirow{2}{*}{\shortstack{\textbf{\name} \\ \textbf{(Finetuned)}}}  & 3 & 36.04 & 12.01 &  \textcolor{blue}{\textbf{65.82}} & 20 & 6.55 &  \textcolor{blue}{\textbf{84.73}} & \textcolor{blue}{\textbf{97.93}} & \textcolor{blue}{\textbf{89.63}} & \textcolor{blue}{\textbf{76.96}}\\[2.2ex]
\bottomrule
\end{tabular}
\end{table*}

\begin{figure*}[!ht]
    \centering

    \subfloat[Comm. Speed-up for CIFAR-10]{%
        \includegraphics[width=0.32\textwidth]{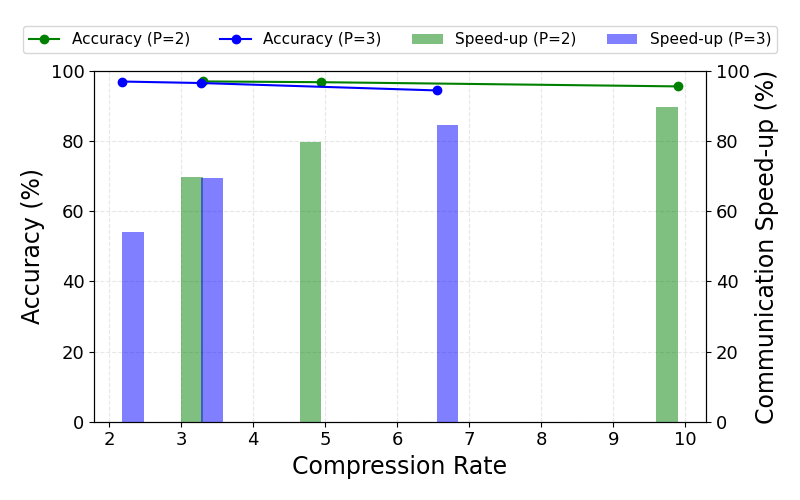}%
        \label{fig:accVSComm_CIFAR10}}%
    \hfill
    \subfloat[Comm. Speed-up for CIFAR-100]{%
        \includegraphics[width=0.32\textwidth]{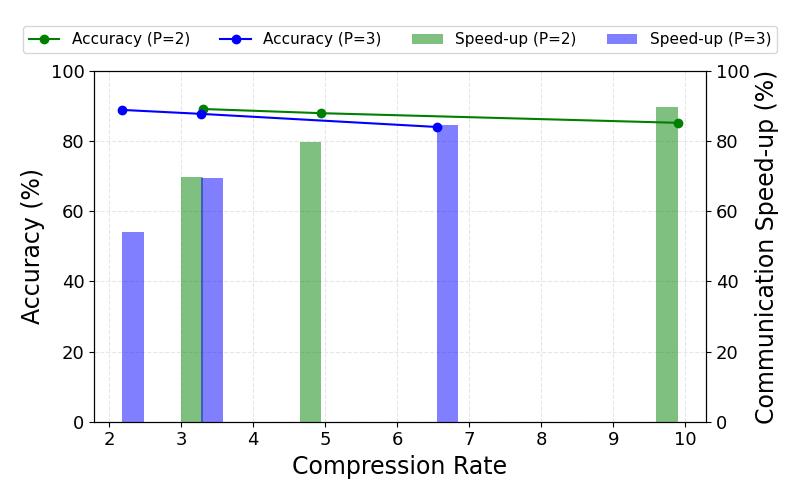}%
        \label{fig:accVSComm_CIFAR100}}%
    \hfill
    \subfloat[Comm. Speed-up for ImageNet1k]{%
        \includegraphics[width=0.32\textwidth]{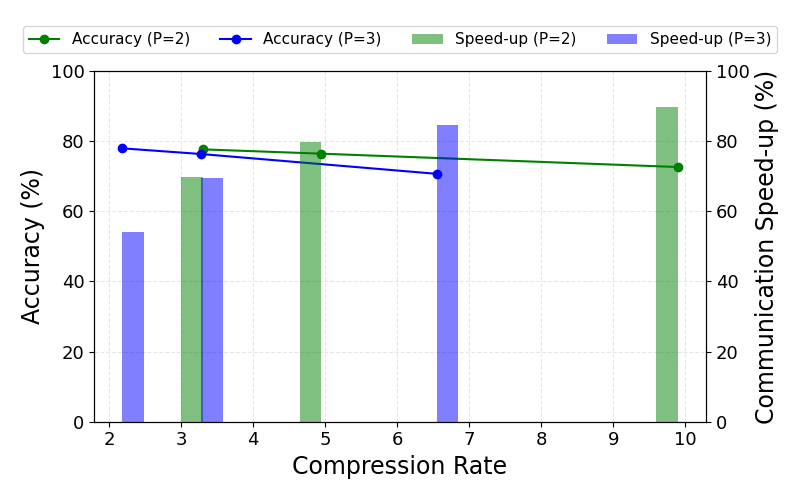}%
        \label{fig:accVSComm_ImageNet}}%

        \caption{Impact of compression rate on accuracy and speed-up for ViT evaluated on CIFAR-10, CIFAR-100 and ImageNet-1k datasets, using partitioning schemes with $p=2$ and $p=3$. Solid lines correspond to model accuracy (left y-axis), while bars indicate speed-up (right y-axis). (a-c) illustrates the trade-off between accuracy and computation speed-up. 
                }
    \label{fig:acc_vs_speedup}
\end{figure*}

\begin{figure*} [!t]
\centering
    \centering
    \subfloat[$P=2$]{%
        \includegraphics[width=0.49\textwidth]{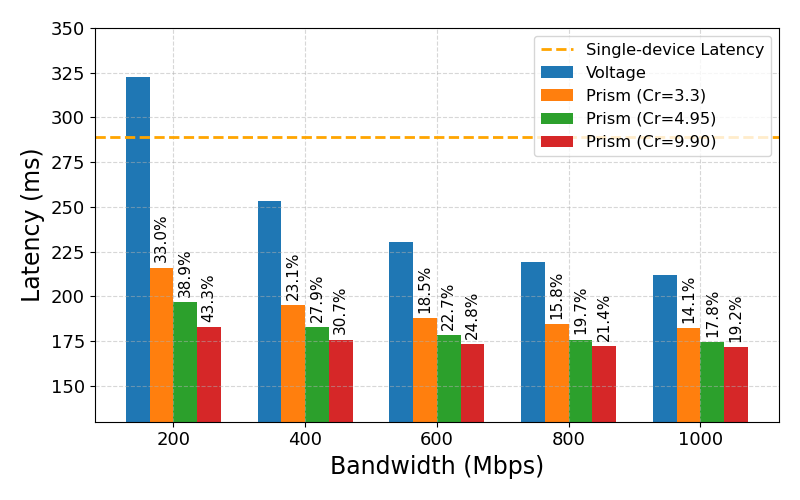}%
        \label{fig:model-a}}%
    \hfill
    \subfloat[$P=3$]{%
        \includegraphics[width=0.49\textwidth]{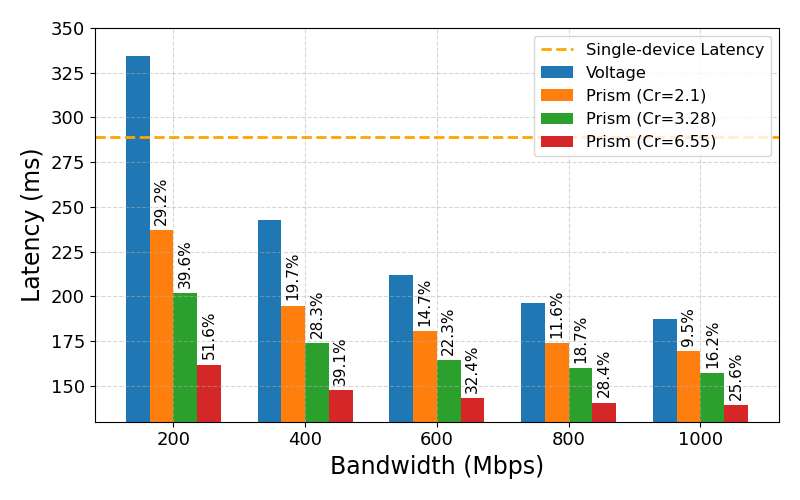}%
        \label{fig:model-b}}%

    \vspace{-0.5em}
\caption{Latency comparison across bandwidths for the ViT model. The orange line indicates the latency of a single-device deployment. Increasing the compression rate significantly reduces latency.}
\label{fig:latency-subfigs}
\end{figure*}

\begin{table*}[!htbp]
\centering
\caption{Comparison of Computation and Communication Efficiency For BERT Model}
\label{tab:bert_strategy_comparison}
\setlength{\tabcolsep}{1pt} 
\begin{tabular}{lm{0.3cm}m{1.2cm}m{1.2cm}m{1.4cm}m{1.3cm}m{0.6cm}m{1.3cm}m{1.0cm}m{0.9cm}m{0.9cm}m{0.9cm}m{0.8cm}m{0.9cm}m{1cm}m{0.8cm}}
\toprule
\multirow{2}{*}{\textbf{Strategy}} & \multirow{2}{*}{\textbf{P}} & \textbf{GFLOPs} & \textbf{GFLOPs} & \textbf{Comp. Speed-up} & \textbf{PDPLC} & \multirow{2}{*}{\textbf{CR}}&
\textbf{Comm. Speed-up} & \textbf{STS-B} & \textbf{SST-2} & \textbf{RTE} & \textbf{QQP} & \textbf{QNLI} & \textbf{MRPC} & \textbf{COLA} & \textbf{MNLI} \\
\cline{3-6}\cline{8-16}
&&\textbf{Total}& \textbf{/device}&\textbf{\%}&\textbf{Tokens}&&\textbf{\%}&\textbf{Spear-man}&\textbf{Acc.}&\textbf{Acc.}&\textbf{F1}&\textbf{Acc.}&\textbf{F1}&\textbf{MCC}&\textbf{Acc.}\\
\midrule
No partition & 1 & 45.93 & 45.93 & - & - & - & - & 88.12 & 92.77 & 67.50 & 88.11 & 91.48 & 87.39 & 56.11 & 84.70 \\
Voltage \cite{10631032} & 2 & 53.18 & 26.59 & 42.11 & 128 & - & - & 88.12 & 92.77 & 67.50 & 88.11 & 91.48 & 87.39 & 56.11 & 84.70 \\
Voltage \cite{10631032} & 3 & 60.42 & 20.14 & 56.15 & 171 & - & - & 88.12 & 92.77 & 67.50 & 88.11 & 91.48 & 87.39 & 56.11 & 84.70 \\
\midrule
\multirow{4}{*}{\textbf{\name}} 
  & 2 & 45.58 & 22.79 & 50.38 & 13 & 9.85 & 89.84 & \textcolor{blue}{\textbf{88.12}} & \textcolor{blue}{\textbf{92.77}} & \textcolor{blue}{\textbf{65.70}} & \textcolor{blue}{\textbf{88.11}} & \textcolor{blue}{\textbf{91.48}} & \textcolor{blue}{\textbf{87.39}} & \textcolor{blue}{\textbf{56.11}} & \textcolor{blue}{\textbf{84.59}} \\
  & 2 & 44.79 & 22.40 & 51.24 & 1 & 128 & \textcolor{blue}{\textbf{99.22}} & \textcolor{blue}{\textbf{88.12}} & \textcolor{blue}{\textbf{92.77}} & \textcolor{blue}{\textbf{65.70}} & \textcolor{blue}{\textbf{88.11}} & \textcolor{blue}{\textbf{91.48}} & \textcolor{blue}{\textbf{87.39}} & \textcolor{blue}{\textbf{56.11}} & 84.53 \\
  & 3 & 46.02 & 15.34 & 66.60 & 18 & 9.50 & 89.47 & \textcolor{blue}{\textbf{88.12}} & \textcolor{blue}{\textbf{92.77}} & 62.81 & \textcolor{blue}{\textbf{88.11}} & \textcolor{blue}{\textbf{91.48}} & \textcolor{blue}{\textbf{87.39}} & \textcolor{blue}{\textbf{56.11}} & 84.19 \\
  & 3 & 44.51 & 14.84 &  \textcolor{blue}{\textbf{67.70}} & 2 &85.50 & 98.83 & \textcolor{blue}{\textbf{88.12}} & \textcolor{blue}{\textbf{92.77}} & 62.45 & \textcolor{blue}{\textbf{88.11}} & \textcolor{blue}{\textbf{91.48}} & \textcolor{blue}{\textbf{87.39}} & \textcolor{blue}{\textbf{56.11}} & 84.11 \\
\bottomrule
\end{tabular}
\end{table*}
\begin{table*}[!th]
\centering
\caption{Comparison of Computation and Communication Efficiency For GPT-2 Model}
\label{tab:gpt_strategy_comparison}
\begin{tabular}{lm{1.00cm}m{1.2cm}m{1.2cm}m{1.2cm}m{1cm}m{1.3cm}m{1.2cm}m{1.2cm}m{1cm}m{1cm}}
\toprule
\multirow{2}{*}{\textbf{Strategy}} & \multirow{2}{*}{\textbf{P}} & \textbf{GFLOPs} & \textbf{GFLOPs} & \textbf{Comp. Speed-up} & \multirow{2}{*}{\textbf{CR}} & \textbf{Comm. Speed-up} & \textbf{CBT-CN} & \textbf{CBT-NE} & \textbf{Enwik8} & \textbf{Text8} \\
\cline{3-5}\cline{7-11}
&&\textbf{Total}& \textbf{/device}&\textbf{\%}&&\textbf{\%}&Acc.&Acc.&BPB&BPC \\
\midrule
No partition & 1 & 65.71 & 65.71 & -- & 1 & -- & 79 & 80 & 1.34 & 1.21 \\
Voltage \cite{10631032} & 2 & 72.97 & 36.49 & 44.48 & 1 & -- & 79 & 80 & 1.34 & 1.21 \\
Voltage \cite{10631032} & 3 & 80.23 & 26.74 & 59.30 & 1 & -- & 79 & 80 & 1.34 & 1.21 \\
\midrule
\multirow{18}{*}{\textbf{\name}} 
& 2 & 68.71 & 34.36 & 47.72 & 2 & 50.00 & \textcolor{blue}{\textbf{76}} & \textcolor{blue}{\textbf{76}} & \textcolor{blue}{\textbf{1.39}} & \textcolor{blue}{\textbf{1.23}} \\
& 2 & 67.26 & 33.63 & 48.82 & 3 & 66.67 & 75 & 74 & 1.41 & 1.24 \\
& 2 & 66.60 & 33.30 & 49.32 & 4 & 75.00 & 75 & 75 & 1.42 & 1.24 \\
& 2 & 66.13 & 33.07 & 49.68 & 5 & 80.00 & 74 & 74 & 1.43 & 1.25 \\
& 2 & 65.87 & 32.94 & 49.88 & 6 & 83.33 & 74 & 74 & 1.44 & 1.26 \\
& 2 & 65.67 & 32.84 & 50.03 & 7 & 85.71 & 73 & 74 & 1.44 & 1.26 \\
& 2 & 65.54 & 32.77 & 50.13 & 8 & 87.50 & 73 & 74 & 1.45 & 1.27 \\
& 2 & 65.41 & 32.71 & 50.23 & 9 & 88.89 & 72 & 73 & 1.45 & 1.28 \\
& 2 & 65.27 & 32.64 &  \textcolor{blue}{\textbf{50.33}} & 10 & \textcolor{blue}{\textbf{90.00}} & 72 & 72 & 1.46 & 1.29 \\
\cline{2-11}
& 3 & 72.02 & 24.01 & 63.47 & 2 & 50.00 & 75 & 73 & 1.42 & 1.24 \\
& 3 & 69.37 & 23.12 & 64.81 & 3 & 66.67 & 74 & 70 & 1.46 & 1.26 \\
& 3 & 68.05 & 22.68 & 65.48 & 4 & 75.00 & 73 & 70 & 1.48 & 1.26 \\
& 3 & 67.29 & 22.43 & 65.87 & 5 & 80.00 & 72 & 69 & 1.49 & 1.27 \\
& 3 & 66.72 & 22.24 & 66.15 & 6 & 83.33 & 72 & 69 & 1.50 & 1.28 \\
& 3 & 66.35 & 22.12 & 66.34 & 7 & 85.71 & 71 & 68 & 1.51 & 1.29 \\
& 3 & 65.97 & 21.99 & 66.53 & 8 & 87.50 & 71 & 68 & 1.52 & 1.30 \\
& 3 & 65.78 & 21.93 & 66.63 & 9 & 88.89 & 70 & 68 & 1.53 & 1.31 \\
& 3 & 65.59 & 21.86 &  \textcolor{blue}{\textbf{66.73}} & 10 & \textcolor{blue}{\textbf{90.00}} & 70 & 67 & 1.53 & 1.32 \\
\bottomrule
\end{tabular}
\end{table*}

\subsection{Overall Performance}    

We begin our performance analysis by evaluating the effectiveness of \name on the ViT model using CIFAR-10, CIFAR-100, and ImageNet-1K datasets. As summarized in Table~\ref{tab:strategy_comparison}, \name achieves substantial communication speed-up across all configurations. Specifically, with $P = 2$ participating devices and a compression rate of $\textit{CR} = 9.9$, \name delivers an 89.90\% communication speed-up. Similarly, for the case of $P = 3$ and $\textit{CR} = 6.55$, it yields 84.73\% communication speed-up. It is worth noting that, while data exchange among edge devices can be realized through broadcast when $P >2$, we assume unicast data transmission in the comparison in order to follow the same assumption used in Voltage~\cite{10631032}. In other words, the broadcast transmission can further speed up communication. In terms of computation, a higher computation speed-up is observed with $P = 3$, where per-device GFLOPs decrease by 65.8\% compared to the single-device baseline setup, highlighting the scalability and efficiency of \name for distributed edge inference. We also analyze the trade-offs between compression rate, communication speed-up, and model accuracy. As shown in Fig.~\ref{fig:acc_vs_speedup}, increasing the compression rate contributes to higher communication speed-up but causes more accuracy loss. For instance, at $\textit{CR} = 6.55$ with $P = 3$, the communication speed-up reaches $84.73\%$. However, model accuracy drops from $98.01\%$ to $94.49\%$ on CIFAR-10, from $91.00\%$ to $84.07\%$ on CIFAR-100, and from $80.30\%$ to $70.70\%$ on ImageNet-1K. However, after fine-tuning the model under the same settings, accuracy improves to $97.93\%$, $89.63\%$, and $76.96\%$ on CIFAR-10, CIFAR-100, and ImageNet-1K, respectively. It is also observed that for similar \textit{CR} values, the $P = 3$ setup tends to exhibit slightly high accuracy drops than $P = 2$. For example, at \textit{CR} $\approx$ 3.3, accuracy for CIFAR-100 is 89.2\% with $P = 2$ and 88.9\% with $P = 3$. This is reasonable, as for a larger output feature, its \sm contain more context information at the same \textit{CR}. The insight indicates that it is important to choose an appropriate \textit{CR} to strike a balance on communication overhead reduction and accuracy, depending on the size of intermediate output feature.

To realistically simulate edge deployments, all latency evaluations were conducted on a CPU setup with 2 CPU cores running at 2.1\,GHz. We set the input data batch size to 1 and used an input sample size of $224 \times 224 \times 3$ for the ViT model. As shown in Fig.~\ref{fig:latency-subfigs}, \name consistently outperforms the Voltage~\cite{10631032} method across all bandwidths. Even at 200\,Mbps, \name reduces latency by 43.3\% for $P = 2$ with $\textit{CR} = 9.9$, and by 52.6\% for $P = 3$ with $\textit{CR} = 6.55$, whereas Voltage performs worse than the single-device baseline setup. Furthermore, as the bandwidth increases, while the latency decreases for both \name and Voltage, the relative advantage of \name remains strong under all network conditions. 

Table~\ref{tab:bert_strategy_comparison} presents the results of applying \name to the BERT model on a suite of GLUE benchmark tasks. At $P = 3$ and a compression rate of \textit{CR} = 85.50, per-device computation is reduced by 67.7\% and communication overhead is reduced by 98.83\%. Even with such an aggressive compression, BERT's performance remains stable across most tasks. This is mainly due to the smaller number of dataset classes compared to the datasets in ViT. For instance, \name achieves STS-B, SST-2, QQP, QNLI, and MRPC scores as those of the original model, while RTE and MNLI see a slight drop. Notably, \name achieves similar communication reduction (99.2\%) even with $P = 2$, while keeping accuracy virtually unchanged. This observation suggests that for classification tasks with smaller number of classes, a higher compression of \sm can be applied without significant effect on the model accuracy.

For GPT-2, Table~\ref{tab:gpt_strategy_comparison} shows that with $P = 3$ and \textit{CR} = 10, \name achieves a 66.7\% reduction in per-device computation and 90\% reduction in communication overhead. Accuracy degrades slightly on CBT and byte-level tasks at higher compression, but remains within acceptable limits. For example, BPC on Text8 increases from 1.21 (baseline) to 1.32 at the most aggressive compression setting, and can be partially recovered through fine-tuning.

These results highlight the broad applicability of \name across different Transformer architectures and tasks—from image classification to sentence classification and autoregressive language modeling. In all cases, \name substantially reduces communication overhead with minor or no impact on model accuracy. Furthermore, for applications requiring higher accuracy under aggressive compression, performance can be improved by fine-tuning the model or by training it from scratch with \name integrated into the training pipeline from the beginning.

\section{Conclusion} \label{conclusion}

In this work, we presented \name, an efficient and scalable strategy for distributed inference of Transformer-based models, addressing the key bottlenecks of communication overhead and per-device computation cost. We introduced a \sm-based representation, which compresses intermediate outputs with minor accuracy degradation and an optimized attention mechanism reducing the redundant Key/Value computations. Additionally, we provide a partition-aware causal masking solution that ensures correctness in autoregressive inference. Experimental results using ViT, BERT, and GPT-2 demonstrate that \name significantly reduces communication overhead and computations across diverse tasks and datasets, demonstrating its strong potential for real-time, edge-level, and multi-device deployment scenarios. This approach sets the foundation for future improvements to communication-efficient, distributed Transformer inference. 

\section*{Acknowledgments}
This research was supported by the PANDORA project, funded by the European Union’s Horizon Europe Framework Programme under Grant Agreement No 101135775 and NordForsk Nordic University Cooperation on Edge Intelligence (Grant No. 168043).

\bibliographystyle{IEEEtran}
\bibliography{base}
\end{document}

%% file: mycommands.tex
\usepackage{xspace} 

\newcommand{\name}{\textsc{Prism}\xspace}
\newcommand{\sm}{Segment Means\xspace}